\documentclass[10pt,twocolumn]{article}
\usepackage[utf8]{inputenc}
\usepackage[T1]{fontenc}
\usepackage{mathptmx}
\usepackage[scaled=0.92]{helvet}
\usepackage{courier}
\usepackage[a4paper,margin=2cm,columnsep=0.7cm]{geometry}
\usepackage{amsmath}
\usepackage{graphicx}
\usepackage{tikz}
\usetikzlibrary{arrows.meta,positioning,fit,calc}
\usepackage{booktabs}
\usepackage{array}
\usepackage[font=small,labelfont=bf]{caption}
\usepackage[hidelinks]{hyperref}
\usepackage{url}
\usepackage[section]{placeins}

\title{\textbf{Implementation of Machine Translation for the Uzbek Language Based on Transformer Models}}
\author{Mukhammadjon Musaev\\
\small Tashkent University of Information Technologies\\
\small named after Muhammad al-Khwarizmi, Uzbekistan\\
\small \texttt{mm.musaev@rambler.ru}
\and
Mannon Ochilov\\
\small Tashkent University of Information Technologies\\
\small named after Muhammad al-Khwarizmi, Uzbekistan\\
\small \texttt{ochilov.mannon@mail.ru}}
\date{}

\begin{document}

\twocolumn[
\maketitle
\begin{@twocolumnfalse}
\begin{abstract}
This research focuses on building a machine translation system for the Uzbek language using transformer models such as T5 and mT5. Because of the limited data and the complexity of the Uzbek language, we collect, preprocess and augment data to create a strong Uzbek--English parallel corpus of about 3 million sentence pairs. The models are fine-tuned for optimal translation performance, and we use the BLEU score to evaluate the results. This study shows that our approach significantly improves upon older methods and helps promote the Uzbek language in the digital world.
\end{abstract}
\vspace{1em}
\end{@twocolumnfalse}
]

{\renewcommand{\thefootnote}{}\footnotetext{This is the authors' version of a paper published in the Proceedings of the 1st International Scientific-Practical Conference ``Digital Transformation and Artificial Intelligence: Problems, Innovations and Trends'' (DTAI-2024), Tashkent State University of Economics, Tashkent, 11 September 2024, pp.~450--456. Compared with the published version, the reference list has been corrected and minor typographical errors have been fixed.}}

\section{Introduction}
Machine translation, the process of automatically translating text from one language to another, has advanced significantly with the development of artificial intelligence algorithms. The use of computers for translating text between languages has long been a goal of computer science. However, only in the last ten years has machine translation become an effective tool for wider use. Breakthroughs in natural language processing, artificial intelligence and mathematics have contributed to the creation of increasingly useful technologies in this field.

Machine translation (MT) is the process of automatically translating text from one language to another without human intervention. Since the 1950s, machine translation has been considered one of the first tasks faced by computers. Unfortunately, the complexity of the task turned out to be much greater than initial calculations suggested, requiring enormous processing power far beyond the capabilities of early computers. Only in the early 2000s did software, data and the necessary hardware become capable of performing basic machine translation. The developers of the early machine translation programs used statistical databases of languages to ``teach'' computers to translate text. Training these systems required a significant amount of manual labour: for each newly added language, the work had to start from collecting statistical data, which demanded a lot of time, effort and financial resources~\cite{koehn2010}.

\section{Methodology}
The most common types of machine translation are the following.

\textbf{1. Rule-based MT.} The oldest form of MT, which has several major drawbacks, including the need for extensive human post-editing, the need to add languages manually, and generally low quality. It is used in very simple situations where quick understanding is required~\cite{hutchins1986}.

\textbf{2. Statistical MT.} Statistical MT builds a statistical model of the relationships between words, phrases and sentences in a text~\cite{brown1993}. To translate these elements into a new language, a statistical model must also be built for the second language. It improves upon rule-based MT. Common techniques in statistical MT include sentence-based models and $N$-gram language models. Statistical machine translation (SMT) relies on statistical models to translate text from one language to another. Its key concepts are:
\begin{itemize}\setlength\itemsep{0pt}
\item \emph{Parallel corpus:} SMT relies on parallel corpora, which are collections of texts in two or more languages that are translations of each other. These corpora are used to train statistical models.
\item \emph{Phrase-based models:} one of the core architectures of SMT. Sentences are broken into smaller units called phrases, which are translated independently, and the model learns the translation probabilities for each phrase pair from the parallel corpus~\cite{koehn2003}.
\item \emph{Language models:} in addition to translation probabilities, SMT systems use language models to assess the likelihood of a sequence of words in the target language, improving the fluency of translations.
\item \emph{Alignment model:} SMT systems also include alignment models that learn the correspondences between words or phrases in the source and target languages~\cite{och2003}.
\item \emph{Decoding:} during translation, SMT systems use a decoding algorithm to search for the most likely translation of the input sentence according to the translation and language models.
\end{itemize}
Although SMT was widely used and achieved significant success, it has several limitations. SMT systems struggle with translating phrases, handling word-order differences between languages, and producing fluent and natural translations, especially for languages with complex syntax or morphology. Overall, while SMT has mostly been replaced by neural machine translation (NMT), it remains relevant in certain scenarios, particularly for language pairs with limited resources or specific domain requirements.

\textbf{3. Neural Machine Translation (NMT).} NMT uses artificial neural networks to translate text directly. These systems are more accurate, make it easier to add new languages, and operate faster once trained. NMT typically employs an end-to-end model built on recurrent neural networks (RNN)~\cite{sutskever2014}, convolutional neural networks (CNN)~\cite{gehring2017} or transformers~\cite{vaswani2017}. NMT models can learn complex relationships between the source and target languages, leading to smoother and more accurate translations than SMT~\cite{koehn2017}. Key aspects of NMT are:
\begin{itemize}\setlength\itemsep{0pt}
\item \emph{End-to-end learning:} unlike SMT, which involves several stages such as phrase-based translation and language modelling, NMT learns the mapping from the source language to the target language directly within a single neural network.
\item \emph{Recurrent neural networks:} early NMT models often used RNNs, particularly long short-term memory (LSTM) networks, because of their ability to capture temporal dependencies in sequential data~\cite{hochreiter1997,sutskever2014}.
\item \emph{Attention mechanism:} attention allows the model to focus on different parts of the source sentence when generating each word of the target sentence, which alleviates the problem of long-term dependencies~\cite{bahdanau2015}.
\item \emph{Transformer architecture:} transformers use self-attention to capture global dependencies within input sentences, enabling parallelisation and improving performance on long sequences. They have largely replaced RNN-based models.
\end{itemize}
NMT models can be trained to translate between multiple language pairs at once, and some support zero-shot translation between pairs never seen explicitly during training~\cite{johnson2017}. NMT offers better fluency, improved accuracy and the ability to capture long-distance dependencies, requires less manual feature engineering, and has become the dominant approach in the field.

\textbf{4. Attention mechanisms.} Attention mechanisms are often integrated into NMT architectures, allowing the model to focus on relevant parts of the input sentence during translation~\cite{bahdanau2015}.

\textbf{5. Transformer models.} Transformers, introduced in the paper ``Attention Is All You Need''~\cite{vaswani2017}, have become the core of modern machine translation. They use self-attention to assess the importance of different words in a sentence, enabling parallelisation and capturing long-term dependencies. Models such as BERT~\cite{devlin2019} and GPT~\cite{radford2018} have been adapted for machine translation tasks~\cite{tay2022}.

\textbf{6. Transfer learning.} Transfer learning involves pre-training models on large datasets and then fine-tuning them for specific translation tasks. Pre-trained language models are trained on large datasets and learn contextual representations of language; they can then be fine-tuned on bilingual data~\cite{zoph2016}, updating their parameters while retaining the knowledge gained during pre-training. Encoder--decoder architectures can likewise be pre-trained and fine-tuned on parallel corpora. Transfer learning leads to better initialisation and faster convergence, is useful for low-resource language pairs, and is particularly helpful for domain adaptation (e.g., medical, legal or technical texts). It can also be combined with multi-task learning, for example by jointly training a model on several language pairs or on related tasks such as language modelling~\cite{luong2016}. Its challenges include domain mismatch between pre-training data and the target task and the need for careful selection of hyper-parameters and training strategies.

\textbf{7. Reinforcement learning.} Reinforcement learning methods can be applied to optimise translation quality: the translation model interacts with an environment (e.g., human evaluators) and receives feedback on its translations, which it uses to improve over time~\cite{nguyen2017}.

\textbf{8. Unsupervised machine translation.} Unsupervised MT aims to translate between languages without parallel corpora. Methods such as adversarial training and back-translation have been proposed to train translation models using only monolingual data in each language~\cite{lample2018,sennrich2016,yang2018}.

\textbf{9. Hybrid approaches.} Hybrid approaches combine different methods, such as statistical and neural MT, to leverage their strengths, for example by using neural networks for translation while incorporating linguistic knowledge or rules from statistical models~\cite{koehnknowles2017}.

Developers of machine translation technology such as Google, Microsoft and Amazon currently favour neural MT, because it provides more accurate translations and allows new language pairs to be added continuously~\cite{wu2016}.

\section{Dataset Preparation}
The first step in creating machine learning models for translation into Uzbek is to develop a training dataset, in our case a bilingual parallel corpus. A parallel corpus is a collection of texts in two or more languages aligned at the sentence or phrase level. Parallel corpora are invaluable resources for natural language processing tasks such as machine translation, bilingual dictionary creation, cross-lingual information retrieval and language learning.

In this research we focus on machine translation between Uzbek and English, so the parallel corpus is developed for these two languages. A small sample is shown in Table~\ref{tab:sample}.

\begin{table}[!htbp]
\caption{Examples from the parallel corpus}
\label{tab:sample}
\centering\small
\begin{tabular}{@{}r>{\raggedright\arraybackslash}p{3.3cm}>{\raggedright\arraybackslash}p{3.3cm}@{}}
\toprule
No. & Uzbek & English \\
\midrule
1 & Maning ismim Mubina. & My name is Mubina. \\
2 & Men 20 yoshdaman. & I am 20 years old. \\
3 & Men sun’iy intellekt sohasiga qiziqaman. & I am interested in the field of artificial intelligence. \\
4 & Men matematika, fizika va tabiiy fanlarini yaxshi ko‘raman. & I like mathematics, physics and natural sciences. \\
\ldots & \ldots & \ldots \\
\bottomrule
\end{tabular}
\end{table}

For sentence-level Uzbek--English translation we need a parallel corpus of complete sentences written in Uzbek together with their English counterparts. Punctuation and letter case must also be taken into account, since the case of a word or the placement of a punctuation mark can change the meaning of a sentence.

We perform machine translation for Uzbek by fine-tuning the multilingual text-to-text transfer transformer mT5~\cite{xue2021}, which requires a large parallel corpus. The corpus was built in four ways.

\textbf{Open-source parallel corpora.} Our analysis showed that there are no extensive, cleaned and prepared parallel corpora specifically for Uzbek--English translation. However, Uzbek--English data are available on the OPUS platform~\cite{tiedemann2012}. Among the available sources, the NLLB (No Language Left Behind) collection~\cite{nllb2022}, based on bitexts mined by Meta AI, contains many pairs, and a large amount of Uzbek--English data is also available from Wikipedia translations and the translation system of the Wikimedia Foundation. We found that the NLLB data cover mainly religious topics, so this source alone is not sufficient for general-purpose translation. In total, approximately 2 million pairs were obtained from OPUS.

\textbf{Creating a parallel corpus through synthesis.} Popular machine translation systems such as Google Translate and Yandex Translate translate between Uzbek and English at a quality close to human translation. We therefore translated Uzbek texts from various sources into English with Google Translate and created approximately 1 million Uzbek--English pairs in this way.

\textbf{Synchronising Uzbek and English texts.} This approach also reduces manual labour and consists of the following steps:
\begin{enumerate}\setlength\itemsep{0pt}
\item Collect texts that are available in both Uzbek and English.
\item Split the texts in both languages into smaller segments based on punctuation marks.
\item Translate the Uzbek segments into English with Google Translate (even if an English translation already exists).
\item Measure the similarity between the machine translation and the existing English segments with ready-made tools.
\item For pairs identified as similar, replace the machine translation with the existing English translation.
\item Add the resulting Uzbek--English pairs to the parallel corpus.
\end{enumerate}
Using this approach we synchronised texts published on official Uzbek websites and collected approximately 30,000 pairs.

\textbf{Manual translation.} Since the beginning of the research, approximately 20,000 Uzbek texts have been translated into English manually over the course of two years.

The characteristics of the resulting corpus are given in Table~\ref{tab:corpus}, and its division into training, validation and test sets in Table~\ref{tab:split}.

\begin{table}[!htbp]
\caption{Characteristics of the Uzbek--English parallel corpus}
\label{tab:corpus}
\centering\small
\setlength{\tabcolsep}{4pt}
\begin{tabular}{@{}lrrr@{}}
\toprule
Source & Rows & Sentences & Tokens \\
\midrule
Open sources & 1\,980\,450 & 1\,964\,140 & 15\,814\,120 \\
Synthesised & 1\,054\,121 & 1\,038\,425 & 8\,409\,800 \\
Synchronised & 30\,542 & 30\,424 & 283\,392 \\
Manually collected & 20\,130 & 19\,880 & 185\,940 \\
\midrule
Total & 3\,085\,243 & 3\,052\,869 & 24\,693\,252 \\
\bottomrule
\end{tabular}
\end{table}

\begin{table}[!htbp]
\caption{Distribution of the training data}
\label{tab:split}
\centering\small
\setlength{\tabcolsep}{3pt}
\begin{tabular}{@{}lrrrr@{}}
\toprule
 & Total & Training & Validation & Test \\
\midrule
Rows & 3\,085\,243 & 2\,711\,462 & 342\,540 & 31\,241 \\
Sentences & 3\,052\,869 & 2\,682\,347 & 339\,541 & 30\,981 \\
Tokens & 24\,693\,252 & 21\,673\,422 & 2\,781\,420 & 238\,410 \\
\bottomrule
\end{tabular}
\end{table}

\section{Evaluation Metrics}
Evaluating machine translation systems involves assessing their performance, translation quality, fluency and adequacy. The common methods are the following.

\emph{Human evaluation.} Experts assess machine-translated texts according to predefined criteria such as fluency, adequacy and overall accuracy. Human evaluation is considered the gold standard but requires significant time and resources.

\emph{Automatic evaluation metrics.} These measure system quality without human intervention, typically on a scale from 0 (very poor translation) to 1 (perfect translation).

\textbf{BLEU} (Bilingual Evaluation Understudy)~\cite{papineni2002} is one of the most widely used automatic metrics. It measures the similarity between a candidate translation and one or more reference translations using $N$-gram precision, typically for $N$-grams from unigrams up to four-grams; higher precision means that more $N$-grams of the candidate appear in the references. BLEU scores range from 0 to 1, and a score of 1 means that the candidate matches the reference exactly. BLEU relies only on $N$-gram precision and does not fully capture fluency, grammaticality or semantic accuracy, and it is sensitive to sentence length and vocabulary mismatch, so it is often used together with other metrics:
\begin{equation}
\mathrm{BLEU} = \mathrm{BP}\cdot\exp\Big(\sum_{n=1}^{N} \frac{1}{N}\log p_n\Big),
\end{equation}
where $p_n$ is the $N$-gram precision, $N$ is the maximum $N$-gram length (usually $N=4$), and BP is the brevity penalty:
\begin{equation}
\mathrm{BP} =
\begin{cases}
1, & \text{if } c > r,\\
\exp\!\left(1-\dfrac{r}{c}\right), & \text{if } c \le r,
\end{cases}
\end{equation}
with $c$ the length of the candidate output and $r$ the length of the reference. The precision is
\begin{equation}
p_n = \frac{C_n}{C_m},
\end{equation}
where $C_n$ is the number of $N$-grams of the candidate that appear in the reference and $C_m$ is the total number of $N$-grams in the candidate.

\textbf{ROUGE} (Recall-Oriented Understudy for Gisting Evaluation)~\cite{lin2004} is mainly used to evaluate machine-generated summaries but can be adapted to machine translation by comparing word- or $N$-gram-level similarity between candidate and reference. ROUGE was originally recall-oriented, in contrast to the precision-oriented BLEU; today it is usually reported as an F1-score. Its variants include ROUGE-N ($N$-gram overlap, usually for $N=1\ldots4$), ROUGE-W (weighted word matching) and ROUGE-L (longest common subsequence, which tolerates changes in word order and sentence structure and is useful for assessing fluency and coherence). For each variant the F1-score is the harmonic mean of precision $P$ and recall $R$:
\begin{equation}
F_1 = \frac{2\,P\,R}{P+R}.
\end{equation}

\textbf{TER} (Translation Edit Rate)~\cite{snover2006} measures the number of edits (insertions, deletions, substitutions and shifts) required to transform the candidate translation into the reference translation.

\textbf{METEOR} (Metric for Evaluation of Translation with Explicit Ordering)~\cite{banerjee2005} evaluates exact word matches, unigram precision and recall and, optionally, stemming, synonymy and word order, and can be adjusted to specific translation tasks or languages.

Relying on only one of these metrics may not be sufficient for a comprehensive assessment, so it is advisable to evaluate machine translation models with multiple metrics.

\section{Model and Training}
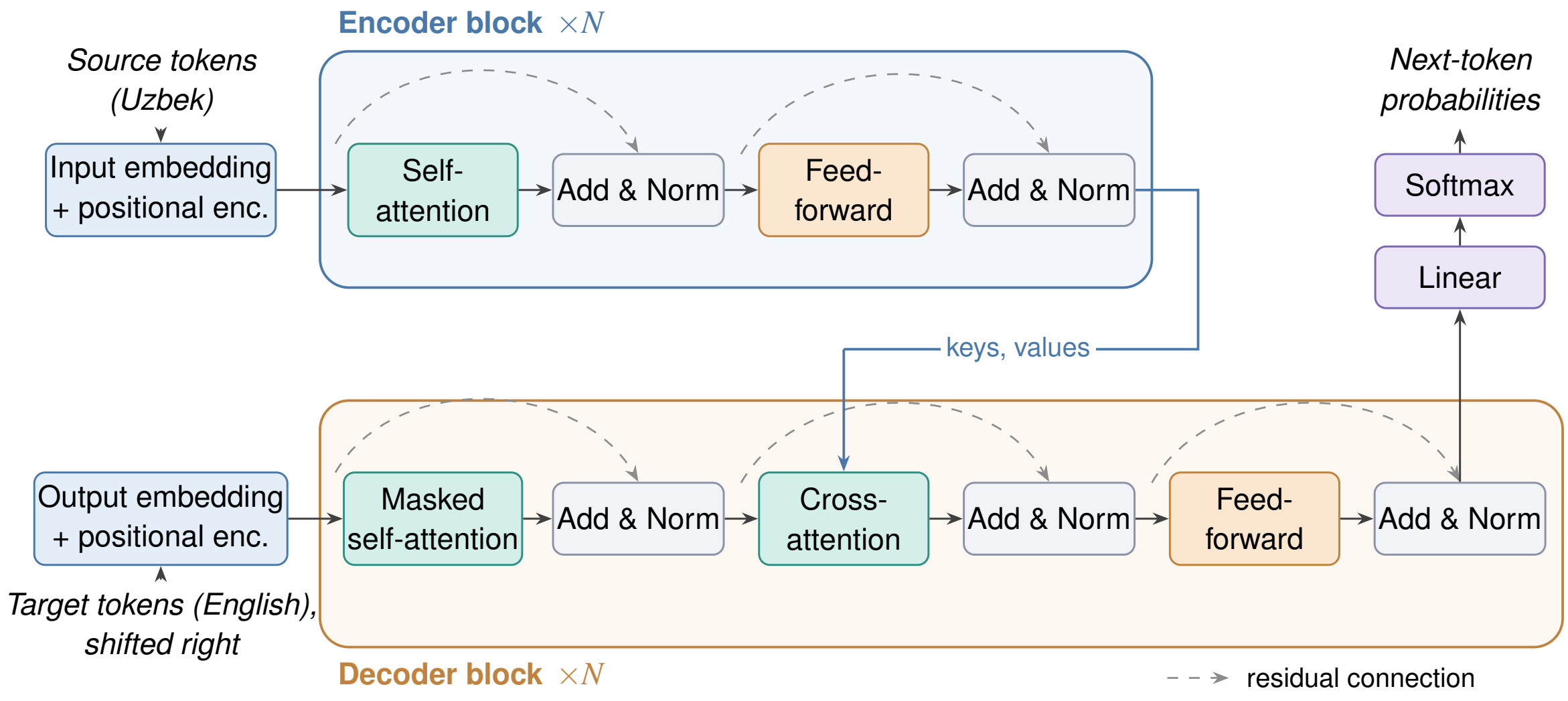
\begin{figure*}[!tbp]
\centering
\resizebox{0.95\textwidth}{!}{
\definecolor{embC}{HTML}{E3EEF9}\definecolor{embB}{HTML}{4A78A8}
\definecolor{attC}{HTML}{D4EEE9}\definecolor{attB}{HTML}{2F8F83}
\definecolor{anC}{HTML}{F2F3F6}\definecolor{anB}{HTML}{8A94A6}
\definecolor{ffC}{HTML}{FBE6D0}\definecolor{ffB}{HTML}{C0833F}
\definecolor{hdC}{HTML}{ECE6F7}\definecolor{hdB}{HTML}{7B68B5}
\definecolor{cardE}{HTML}{F4F8FD}\definecolor{cardD}{HTML}{FDF8F2}\definecolor{frmB}{HTML}{4A78A8}
\begin{tikzpicture}[
  font=\sffamily\small,
  >={Stealth[length=4.5pt]},
  sub/.style={draw, rounded corners=3pt, line width=0.55pt, align=center, minimum width=1.65cm, minimum height=0.9cm, inner sep=1pt},
  att/.style={sub, fill=attC, draw=attB},
  an/.style={sub, fill=anC, draw=anB, minimum height=0.7cm},
  ff/.style={sub, fill=ffC, draw=ffB},
  emb/.style={sub, fill=embC, draw=embB, minimum width=2.1cm},
  hd/.style={sub, fill=hdC, draw=hdB, minimum width=1.65cm, minimum height=0.6cm},
  arr/.style={->, line width=0.55pt, draw=black!75},
  res/.style={->, line width=0.5pt, draw=black!45, dashed},
  card/.style={rounded corners=8pt, line width=0.7pt},
]
\draw[card, fill=cardE, draw=frmB] (3.15,4.25) rectangle (11.25,6.55);
\node[anchor=south west, font=\sffamily\small\bfseries, text=frmB] at (3.2,6.58) {Encoder block \,$\times N$};
\draw[card, fill=cardD, draw=ffB] (3.15,0.75) rectangle (15.25,3.15);
\node[anchor=north west, font=\sffamily\small\bfseries, text=ffB] at (3.2,0.72) {Decoder block \,$\times N$};

\node[font=\sffamily\small\itshape, align=center] (srcT) at (1.6,6.25) {Source tokens\\(Uzbek)};
\node[emb] (iemb) at (1.6,5.2) {Input embedding\\+ positional enc.};
\node[att] (esa)  at (4.25,5.2) {Self-\\attention};
\node[an]  (ean1) at (6.25,5.2) {Add \& Norm};
\node[ff]  (eff)  at (8.25,5.2) {Feed-\\forward};
\node[an]  (ean2) at (10.25,5.2) {Add \& Norm};
\draw[arr] (srcT) -- (iemb);
\draw[arr] (iemb) -- (esa);
\draw[arr] (esa) -- (ean1);
\draw[arr] (ean1) -- (eff);
\draw[arr] (eff) -- (ean2);
\draw[res] (3.3,5.65) to[out=60,in=120] (ean1.north);
\draw[res] (7.25,5.55) to[out=60,in=120] (ean2.north);

\node[emb] (oemb) at (1.6,2.0) {Output embedding\\+ positional enc.};
\node[font=\sffamily\small\itshape, align=center] (tgtT) at (1.6,0.95) {Target tokens (English),\\shifted right};
\node[att] (dsa)  at (4.25,2.0) {Masked\\self-attention};
\node[an]  (dan1) at (6.25,2.0) {Add \& Norm};
\node[att] (dca)  at (8.25,2.0) {Cross-\\attention};
\node[an]  (dan2) at (10.25,2.0) {Add \& Norm};
\node[ff]  (dff)  at (12.25,2.0) {Feed-\\forward};
\node[an]  (dan3) at (14.25,2.0) {Add \& Norm};
\draw[arr] (tgtT) -- (oemb);
\draw[arr] (oemb) -- (dsa);
\draw[arr] (dsa) -- (dan1);
\draw[arr] (dan1) -- (dca);
\draw[arr] (dca) -- (dan2);
\draw[arr] (dan2) -- (dff);
\draw[arr] (dff) -- (dan3);
\draw[res] (3.3,2.45) to[out=60,in=120] (dan1.north);
\draw[res] (7.25,2.35) to[out=60,in=120] (dan2.north);
\draw[res] (11.25,2.35) to[out=60,in=120] (dan3.north);

\draw[line width=0.7pt, draw=frmB] (ean2.east) -- (11.7,5.2) -- (11.7,3.65) -- (8.25,3.65);
\draw[->, line width=0.7pt, draw=frmB] (8.25,3.65) -- (dca.north);
\node[fill=white, inner sep=1.5pt, font=\sffamily\footnotesize, text=frmB] at (9.95,3.65) {keys, values};

\node[hd] (lin) at (14.25,4.35) {Linear};
\node[hd] (sm)  at (14.25,5.25) {Softmax};
\node[font=\sffamily\small\itshape, align=center] (prob) at (14.25,6.25) {Next-token\\probabilities};
\draw[arr] (dan3.north) -- (lin.south);
\draw[arr] (lin) -- (sm);
\draw[arr] (sm) -- (prob);

\begin{scope}[shift={(11.4,0.45)}]
\draw[res] (0,0) -- (0.6,0); \node[anchor=west, font=\sffamily\footnotesize] at (0.65,0) {residual connection};
\end{scope}
\end{tikzpicture}}
\caption{Transformer encoder--decoder architecture used in this work (redrawn after~\cite{vaswani2017}).}
\label{fig:transformer}
\end{figure*}

The Transformer is a deep learning model first proposed in 2017~\cite{vaswani2017}. It employs a self-attention mechanism, which improves performance in neural machine translation and accelerates training compared with recurrent neural networks~\cite{vaswani2017,tay2022}. RNNs can handle inputs of varying length and were once the leading approach for language understanding and machine translation; however, because they compute hidden states sequentially, their parallelisation is limited. The Transformer, by contrast, does not process the input sequentially: self-attention determines the context that gives meaning to each position of the input sequence, which allows greater parallelisation and reduces training time. The Transformer consists mainly of self-attention, an encoder--decoder architecture and feed-forward networks (Fig.~\ref{fig:transformer}).

\textbf{Tokenisation.} In Transformer-based models, text is first converted into tokens and then vectorised: each token is mapped to its index in a vocabulary so that the model can process the text as numeric input. Vocabularies can be built at character, word or subword level. We use subword tokenisation because of its effectiveness for morphologically rich languages such as Uzbek.

While investigating pre-trained multilingual tokenisers on the Hugging Face platform, we found that some already support Uzbek. However, on Uzbek text these tokenisers produce lengthy subword sequences, possibly because of insufficient Uzbek training data. To address this, we created a specialised tokeniser for Uzbek based on the T5Tokenizer~\cite{raffel2020}. We trained a vocabulary of 50,000 tokens with the unigram language model~\cite{kudo2018} using the SentencePiece trainer~\cite{kudorichardson2018}, and added special tokens such as \texttt{PAD}, \texttt{UNK}, \texttt{CLS} and \texttt{SEP}. The unigram model captured the complex characteristics of Uzbek word forms and provided more accurate and efficient tokenisation. An example is shown in Table~\ref{tab:tokens}.

\begin{table}[!htbp]
\caption{Example of tokenisation and vectorisation}
\label{tab:tokens}
\centering\small
\begin{tabular}{@{}l>{\raggedright\arraybackslash}p{5.6cm}@{}}
\toprule
Input text & \emph{Insonlar kitoblarsiz bugungi faravon kunga yetib kelishmas edi.} \\
\midrule
Tokens & [\_Inson, lar, \_kitoblar, siz, \_bugungi, \_faravon, \_kunga, \_yetib, \_kelish, mas, \_edi, ., \texttt{</s>}] \\
\midrule
Token IDs & [7940, 47, 4006, 303, 2453, 41441, 10974, 3517, 5383, 6325, 69, 6, 50100] \\
\bottomrule
\end{tabular}
\end{table}

\begin{figure*}[!tbp]
\centering
\resizebox{0.95\textwidth}{!}{
\definecolor{tealC}{HTML}{D4EEE9}\definecolor{tealB}{HTML}{2F8F83}
\definecolor{sandC}{HTML}{FBE6D0}\definecolor{sandB}{HTML}{C0833F}
\definecolor{steelC}{HTML}{EEF4FB}\definecolor{steelB}{HTML}{4A78A8}
\definecolor{grayC}{HTML}{F2F3F6}\definecolor{grayB}{HTML}{8A94A6}
\definecolor{lilC}{HTML}{ECE6F7}\definecolor{lilB}{HTML}{7B68B5}
\begin{tikzpicture}[
  font=\sffamily\small,
  >={Stealth[length=4.5pt]},
  st/.style={draw, rounded corners=4pt, line width=0.6pt, align=center, inner sep=3pt, minimum height=1.25cm},
  chip/.style={draw=sandB, fill=sandC, rounded corners=2pt, line width=0.5pt, inner sep=2.5pt, font=\sffamily\small\itshape},
  arr/.style={->, line width=0.6pt, draw=black!75},
  lbl/.style={font=\sffamily\footnotesize, text=black!70, fill=white, inner sep=1pt},
]
\node[st, fill=tealC, draw=tealB, text width=2.75cm] (src) at (1.55,3.3) {\textbf{Uzbek input}\\[1pt]\emph{Yoshlar chanada tepadan pastga tushishadi}};
\node[st, fill=grayC, draw=grayB, text width=2.25cm] (tok) at (4.55,3.3) {\textbf{Uzbek tokenizer}\\[1pt]{\footnotesize SentencePiece, unigram, 50k tokens}};

\draw[rounded corners=10pt, line width=0.7pt, draw=steelB, fill=steelC] (6.15,2.25) rectangle (11.55,4.55);
\node[anchor=north west, font=\sffamily\small\bfseries, text=steelB] at (6.25,4.5) {mT5 (fine-tuned)};
\node[st, fill=tealC, draw=tealB, minimum width=1.75cm] (enc) at (7.4,3.15) {\textbf{Encoder}};
\node[st, fill=sandC, draw=sandB, minimum width=1.75cm] (dec) at (10.3,3.15) {\textbf{Decoder}};

\node[st, fill=lilC, draw=lilB, minimum width=1.6cm] (sm) at (12.85,3.15) {Linear +\\Softmax};
\node[chip, draw=lilB, fill=lilC, font=\sffamily\small\bfseries\itshape] (nxt) at (14.6,3.15) {are};
\node[lbl] at (14.6,3.75) {next token};

\draw[arr] (src) -- (tok);
\draw[arr] (tok.east) -- (enc.west|-tok.east);
\draw[arr] (enc) -- (dec);
\node[font=\sffamily\footnotesize, text=black!70] at (8.85,3.95) {hidden states};
\draw[arr] (dec) -- (sm);
\draw[arr] (sm) -- (nxt);

\node[font=\sffamily\small\bfseries, anchor=east] at (7.55,0.85) {Generated English prefix:};
\node[chip] (c1) at (8.2,0.85) {Two};
\node[chip, right=3pt of c1] (c2) {young};
\node[chip, right=3pt of c2] (c3) {people};
\node[chip, right=3pt of c3, dashed, draw=lilB, fill=lilC, font=\sffamily\small\bfseries\itshape] (c4) {are};

\draw[arr] (nxt.south) -- (14.6,0.85) -- (c4.east);
\node[lbl, align=center] at (14.6,1.75) {append and\\repeat until \texttt{</s>}};
\draw[arr] (c3.north) -- ++(0,0.45) -| (dec.south);
\node[lbl] at (11.35,1.55) {decoder input};
\end{tikzpicture}}
\caption{Uzbek--English translation with the fine-tuned mT5 model: the Uzbek sentence is segmented by our Uzbek tokenizer, encoded, and the English translation is generated token by token.}
\label{fig:mt5}
\end{figure*}
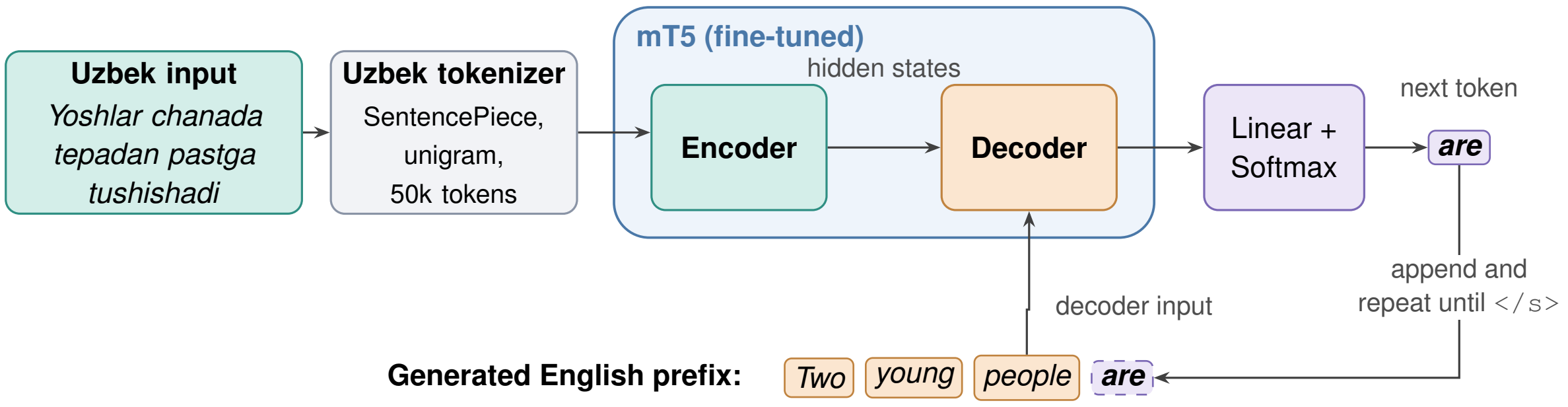

\textbf{Fine-tuning.} We fine-tuned the pre-trained mT5 model on the Uzbek--English dataset described above (Fig.~\ref{fig:mt5}).

\section{Results}
\begin{figure}[!htbp]
\centering
\includegraphics[width=\columnwidth]{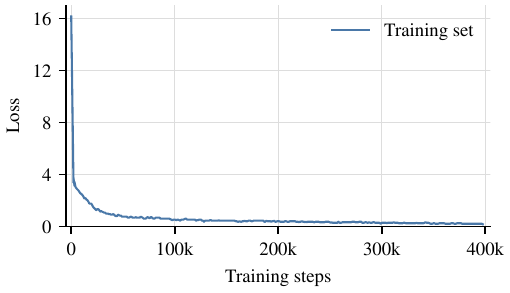}
\caption{Training loss over training steps.}
\label{fig:loss}
\end{figure}

During fine-tuning we used a maximum sequence length of 50, a batch size of 128, a learning rate of $2\times10^{-4}$ and a weight decay of 0.02. Training ran for ten epochs. For optimisation we used the Adafactor optimiser~\cite{shazeer2018} with a learning rate of $5\times10^{-4}$, beta values of 0.9 and 0.999, and an epsilon of $2\times10^{-5}$. The model was trained on an NVIDIA V100S GPU with 32\,GB of memory. The training loss is shown in Fig.~\ref{fig:loss}.

Training took approximately 6 hours per epoch, about 60 hours in total. We evaluated the models with the BLEU metric; the average BLEU (4-gram) scores on the training, validation and test sets are given in Table~\ref{tab:bleu}.

\begin{table}[!htbp]
\caption{BLEU (4-gram) scores of the models}
\label{tab:bleu}
\centering\small
\begin{tabular}{@{}lccc@{}}
\toprule
Model & Training & Validation & Test \\
\midrule
Base model & 0.04 & 0.03 & 0.01 \\
Fine-tuned model & 0.29 & 0.27 & 0.22 \\
Google Translate & 0.37 & 0.36 & 0.34 \\
\bottomrule
\end{tabular}
\end{table}

As Table~\ref{tab:bleu} shows, fine-tuning raises the BLEU (4-gram) score on the test set to 0.22. Examples of translations produced by the fine-tuned model are given in Table~\ref{tab:examples}.

\begin{table}[!htbp]
\caption{Examples of translations produced by the fine-tuned model (BLEU: 1-gram / 4-gram)}
\label{tab:examples}
\centering\scriptsize
\setlength{\tabcolsep}{2.5pt}
\begin{tabular}{@{}r>{\raggedright\arraybackslash}p{2.0cm}>{\raggedright\arraybackslash}p{1.9cm}>{\raggedright\arraybackslash}p{1.7cm}c@{}}
\toprule
No. & Uzbek text & Model translation & Reference & BLEU \\
\midrule
1 & Haftaning juma kunida katta bayram bo‘ladi & There will be a big holiday on Friday of the week & There will be a big holiday on Friday & 0.69 / 0.69 \\
2 & Men talabaman & I am student & I am a student & 0.72 / 0.00 \\
3 & Ob-havo issiq va shamolli. & It’s hot and windy . & It is hot and windy & 0.60 / 0.00 \\
4 & Bu juda yaxshi & It’s very good & This is very good & 0.48 / 0.00 \\
5 & Isming nima? & What is your name & What is your name ? & 0.80 / 0.20 \\
\midrule
\multicolumn{4}{@{}l}{Overall} & 0.66 / 0.18 \\
\bottomrule
\end{tabular}
\end{table}

\section{Conclusion}
In this study, transformer models such as mT5-base and a fine-tuned mT5 were applied to Uzbek--English machine translation. The fine-tuned model achieved a BLEU score of 0.29 on the training set and 0.22 on the test set. On the test set of our dataset, these results are close to those achieved by Google Translate (BLEU 0.34 on the same test set).

\FloatBarrier

\end{document}